\pdfoutput=1

\documentclass[11pt]{article}

\usepackage[]{ACL2023}
\usepackage{graphicx}
\usepackage{times}
\usepackage{latexsym}
\usepackage{lipsum}
\usepackage{hyperref}
\usepackage[T1]{fontenc}

\usepackage[utf8]{inputenc}
\usepackage{booktabs}
\usepackage{microtype}
\usepackage{xspace}
\usepackage{inconsolata}

\newcommand{\repname}{\textsc{DocAMR}\xspace}
\newcommand{\evlname}{\textsc{DocSmatch}\xspace}
\newcommand{\parname}{\textsc{DocParser}\xspace}
\newcommand{\smatch}{\textsc{Smatch}\xspace}

\title{Slide, Constrain, Parse, Repeat: \\ Synchronous Sliding Windows for Document AMR Parsing}

\author{Sadhana Kumaravel \thanks {Co-first author} , 
        Tahira Naseem \footnotemark[1] , 
        Ramon Fernandez Astudillo,\\
        \textbf{Radu Florian, Salim Roukos}\\
        \texttt{\{tnaseem,raduf,roukos\}@us.ibm.com}\\
        \texttt{\{sadhana.kumaravel1,ramon.astudillo\}@ibm.com}\\
        IBM Research AI
        }

\begin{document}
\maketitle
\begin{abstract}
The sliding window approach provides an elegant way to handle contexts of sizes larger than the Transformer's input window, for tasks like language modeling. Here we extend this approach to the sequence-to-sequence task of document parsing. For this, we exploit recent progress in transition-based parsing to implement a parser with synchronous sliding windows over source and target. We develop an oracle and a parser for document-level AMR by expanding on Structured-BART such that it leverages source-target alignments and constrains decoding to guarantee synchronicity and consistency across overlapping windows. We evaluate our oracle and parser using the Abstract Meaning Representation (AMR) parsing 3.0 corpus. On the Multi-Sentence development set of AMR 3.0, we show that our transition oracle loses only 8\% of the gold cross-sentential links despite using a sliding window. In practice, this approach also results in a high-quality document-level parser with manageable memory requirements. Our proposed system \footnote{Code and model available here \href{https://github.com/IBM/transition-amr-parser}{https://github.com/IBM/transition-amr-parser}} performs on par with the state-of-the-art pipeline approach for document-level AMR parsing task on Multi-Sentence AMR 3.0 corpus while maintaining sentence-level parsing performance.


\end{abstract}

\section{Introduction}

Pre-trained Transformer embedding representations \cite{devlin2018bert,liu2019roberta} have shown unparalleled predictive power at sentence-level, but are not built for document encoding and are also not interpretable. Symbolic representations such as Abstract Meaning Representation (AMR) parses \cite{banarescu-etal-2013-abstract}, explicitly consider document-structure \cite{ogorman-etal-2018-amr} and human-understandable concepts, e.g. entities, subject-object relations and actions, and thus can be expected to be complementary to sentence Transformer embedding representations.


Document-level parsing involves both long texts and parses. When only the source (text) or the target (parse) is long, a single sliding window approach is a simple and effective solution \cite{dai-etal-2019-transformer,zaheer2020big,Beltagy2020Longformer}. However, when both have long sizes, the sliding windows must be synchronous, i.e.  words and parse in the overlapping\footnotemark\footnotetext{Note that parsing requires overlap to predict cross-window relations.} part of the source and target windows should correspond to each other. The black-box nature of Transformers prevents us from determining this mapping in a simple manner, precluding the application of sliding windows.

Here we propose an effective solution to apply this approach utilizing recent progress in transition-based parsing. Structured-BART \cite{zhou-etal-2021-structure} builds the parse graph incrementally and aligns the input tokens with parse nodes. The alignment information allows us to make the sliding window on the target side synchronous with the source window. Further, we can constrain the decoding on the overlapping part of the window to obey the previous parse while utilizing position embeddings consistent with the current window. Combining these two observations, we build a document-level Structured-BART parser and oracle.

Our oracle shows that a small window is enough to retain $92\%$ of gold inter-sentential relations\footnote{This statistic was computed on the double-annotation portion of the MS-AMR corpus} in the Multi-Sentence AMR Dataset (MS-AMR) \cite{ogorman-etal-2018-amr}. Further, we match the previous doc-AMR state-of-the-art pipeline \cite{naseem2021docamr} while also matching Structured-BART on the sentence level. To our knowledge, this is also the first parser to apply synchronous sliding windows for document-level parsing, an approach that seems both simple and effective.

\section{Synchronous Sliding Window Parser}

Given the quadratic cost of the attention mechanism in Transformers, many efficient versions have been proposed \cite{gu2021efficiently}. \cite{tay2022efficient,tay2020long} presents a review of such Transformers and their performance on long-range tasks. One emerging pattern is the use of a sliding window, which provides an ingenious method to handle long-range. Previous works leveraging a sliding window \cite{zaheer2020big, Beltagy2020Longformer}, focus on modeling long-range encoder context, while the decoder stays unaware of its own history \cite{ivgi2022efficient}. Modeling the decoder history is particularly important for document-parsing since target history is not just a source of contextual information for the future generation -- it is in fact needed in its full detail for 1) making long-distance co-reference links and 2) keeping the future structurally consistent with the past. 

Below, we present our transition-based parsing system (\textsc{DocParser}) with an encoder-decoder transformer architecture able to use a sliding window both in source and target. Our parser is a modification of the transition-based Structured-BART \cite{zhou-etal-2021-structure}. This model fine-tunes a pre-trained sequence-to-sequence transformer to imitate the actions of an oracle given an input text. These actions, when executed into a state machine, processes the sentence left-to-right and incrementally predict the target graph. 
We first discuss how we adapt the sentence-level state transition system for document-level parsing (section \ref{sec:oracle}) and then move on to the core of our proposal -- the idea of synchronous source and target side sliding windows with constrained decoding (section \ref{sec:slide}).


\subsection{Modified State Machine and Oracle} \label{sec:oracle}

We utilize the state transition system and oracle defined in \newcite{zhou-etal-2021-structure} with modifications to handle inter-sentence relations. The oracle traverses the input tokens from left to write performing one of the following actions at each step:
\begin{quote}
\noindent \textbf{\textsc{shift}:} move cursor to the next token. \\
\noindent\textbf{\textsc{<node>}/\textsc{copy}:} create a node or copy input token under the cursor as node. \\
\noindent\textbf{$\textsc{[la/ra]}({\textup{j},\textsc{lbl}})$:} create an arc with label $\textsc{lbl}$ from the most recent node to the node at the $j_{th}$ step for $\textsc{la}$ and reversed for $\textsc{ra}$. \\
\noindent\textbf{\textsc{root}:} declare the most recent node the root. This can happen once in a sentence. \\
\noindent\textbf{\textsc{close}:} close the state machine; attach unconnected components to the root.
\end{quote} 

Note, that the above state machine can handle only pairwise relations. However, coreference chains in the MS-AMR corpus are annotated over clusters of nodes. To model these chains, we break them into pairwise links. Another aspect that differs in document-level parsing is the notion of the root node. Document graphs do not have a semantic or syntactic root, they only provide relations between a set of sentence-level graphs. In fact, the root node in \repname is a dummy node connecting to all sentence-level roots. Considering this, we modify the state transition system in two ways:

\begin{enumerate}
    \item We introduce a \textsc{close\_sentence} action that allows to relax the one \textsc{root} constraint to one \textsc{root} per sentence. More importantly, any unconnected sub-graphs get attached to the sentence-level root, not the overall graph toot. 
    \item We introduce a special \textsc{same-as} relation label that connects consecutive node pairs belonging to the same coreference chain. 
    The proximity of the chain nodes is determined by the distance between their aligned tokens. Graph node to tokens alignments are created using the approach given in \cite{naseem-etal-2019-rewarding,emnlp2020stacktransformer}.
\end{enumerate}

With these changes, the state machine, along with the original alignment-based oracle of \newcite{zhou-etal-2021-structure}, can seamlessly handle both document and sentence-level AMR graphs. Any actions performed by this state machine are associated with the source token under the cursor; furthermore, the tokens are visited in order from left to right. These two properties come in handy when implementing the sliding window and constrained decoding explained in the next section.

\begin{figure*}[t]
\includegraphics[width=\textwidth]{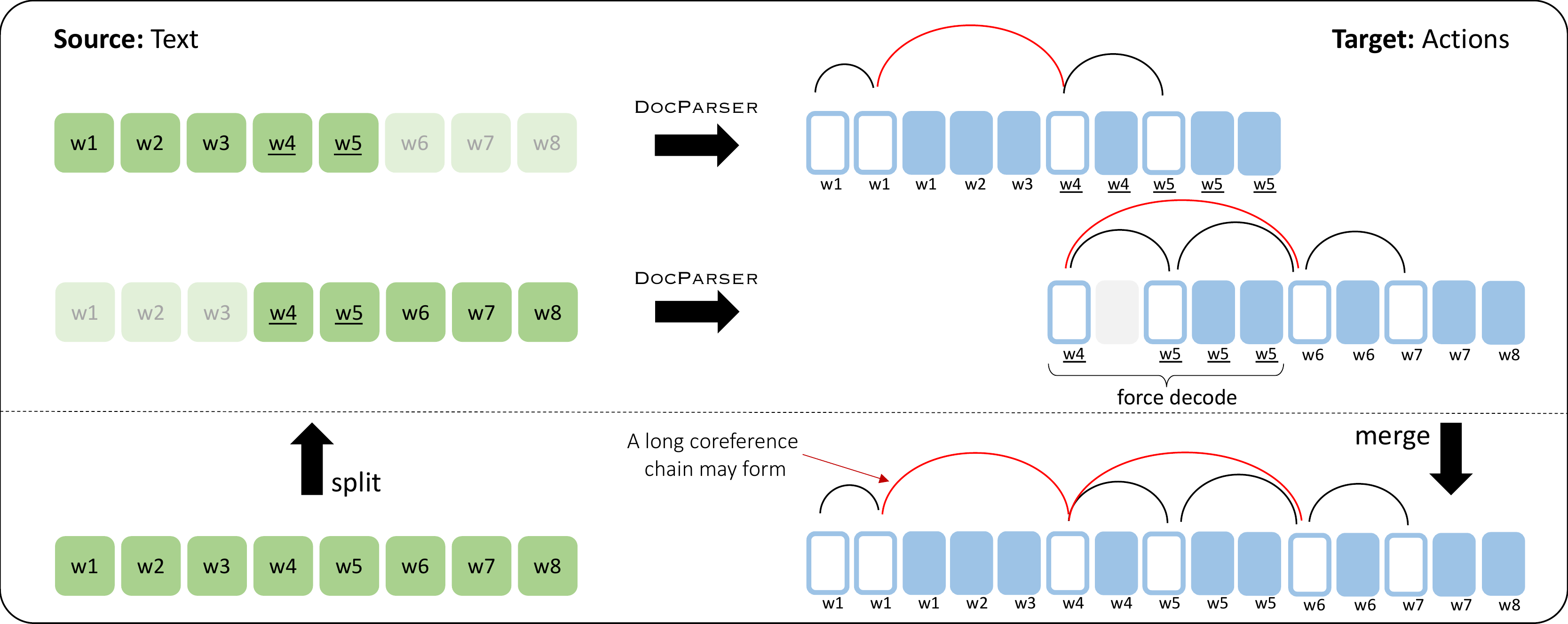}
\caption{Sliding window mechanism for \parname. Green squares (left) represent input tokens and blue rectangles (right) represent actions. Node-making actions have white fill. Actions have token alignments that readily decide which actions to copy over to the next window. Any actions that refer to the past beyond the overlap are not copied over (Greyed-out). After merging, the actions can produce long-ranging co-reference chains.}
\end{figure*} \label{fig:sliding}

\subsection{Sliding Window and Constrained Decoding} \label{sec:slide} 

We base our synchronous parsing approach on the structure-aware BART model of \newcite{zhou-etal-2021-structure} to learn this task. The major modifications to the original BART architecture are a specialized hard cross-attention head \cite{aharoni-goldberg-2017-morphological} to align input tokens to nodes and a dedicated decoder self-attention pointer network \cite{vinyals2015pointer}. This model (similar to other encoder-decoder transformers) can handle source/target lengths of up to 1024 word pieces. The MS-AMR training data has documents that are up to 4277 tokens with an average length of 540 tokens. The source token length typically translates into a target action sequence about 3 times longer -- thus, inputs longer than 300 tokens become unmanageable.  

To handle long documents, we chunk the input text into overlapping windows, the window boundary is aligned with the sentence boundary whenever possible. We parse one window at a time. Each subsequent window must agree with its preceding window regarding the parse of the overlapping portion. This requirement can only be met if we know which parts of the parse belong to the overlapping text.
Our transition-based model incrementally builds the AMR parse by executing transition actions -- these actions are monotonically aligned to the source tokens. 
This makes it easy to keep the graphs of shared portions coherent. We just have to copy over the actions aligned with the overlapping part to the next window and then force decode that portion. When copying over the actions, any connections to the non-overlapping parts of the first window need to be removed and the pointer values in \textsc{la} and \textsc{ra} actions need to be adjusted according to the new action positions. 


Overlapping forced prefixes not only provides a smooth transition between consecutive windows but also makes cross-window links possible -- the actions in the non-forced region of a window (following the prefix) are free to create attachments with the nodes in the forced prefix\footnote{An alternative to force decoding the shared portion could be to repeat the decoder block from the overlapping part as a pre-computed prefix for the next window. This would be more efficient in practice but will make the positional embedding information inconsistent.}. Figure \ref{fig:sliding} shows how the sliding window approach works.

\begin{table}[h]
\begin{center}
\begin{small}
\begin{tabular}{l|c|c|c}
\toprule
& & \multicolumn{2}{c}{DocSmatch (Coref)} \\
\midrule
Window & Overlap & Oracle & \parname \\
\midrule
\midrule
 400 & 200 &  97.8 (92.3) & 64.3 (38)\\
 400 & 100 &  97.6 (91.5) & 64.4 (39)\\
 300 & 200 &  97.7 (91.8) & 65.1 (38)\\
 300 & 100 &  97.4 (90.6) & 65.1 (38)\\
 200 & 100 &  96.7 (87.7) & -- \\
 \midrule
\multicolumn{3}{c|}{\parname w/o sliding decoding} &52.35\\
\bottomrule
\end{tabular}
\end{small}
\end{center}
\caption{Oracle and \parname Smatch and coreference scores with various window and overlap sizes on the double-annotation portion (dev) of MS-AMR 3.0 corpus.
}
    \label{tab:windows}

\end{table}

\begin{table*}[t]
\begin{center}
\begin{tabular}{l|c||c|c||c}
\toprule
 & & \multicolumn{2}{c||}{Multi-Sentence AMR 3.0} & Sentence AMR 3.0  \\
\midrule
System & Train Data & Smatch & DocSmatch (Coref) & Smatch \\
 \midrule
 \midrule
\cite{naseem2021docamr} & \textsc{amr\small{3.0}}\textsc{+conll} & 71.3 & 72.0 (50) & -- \\
\cite{zhou-etal-2021-structure} & \textsc{amr\small{3.0}} & --&-- & 82.3 \\
\midrule
\parname & \textsc{ms-amr} & 65.9 & 66.3 (45) & 71.6 \\
\parname & \textsc{+amr\small{3.0}}\textsc{+conll} & 71.8 & 72.1 (50) & 82.3 \\
\bottomrule
\end{tabular}
\end{center}
\caption{Comparison of \parname with the previous SoTA pipeline approach \cite{naseem2021docamr} on the test set of the document level MS-AMR corpus and with \cite{zhou-etal-2021-structure} on the AMR3.0 sentence level test set.
}
    \label{tab:results}
\end{table*}


\paragraph{What falls off the window?} While the sliding window approach makes long text parsing possible, it has the potential to lose long-range inter-sentence connections. As mentioned in section \ref{sec:oracle}, we break coreference chains into pairwise links between consecutive participant nodes. The full chain can be reconstructed by putting together the pairwise links. To assess the amount of cross-sentence information lost due to window size restriction, we analyze the double-annotation devset of MS-AMR corpus at different window sizes. We first convert document-level AMR annotation into pairwise edges. Then for a given window size and window overlap value, we only retain those gold pairwise edges that are seen in the same window when sliding over the document. We then combine the pairwise edges to reconstruct the document-level AMR graph. We convert this trimmed graph as well as the original gold graph into DocAMR representation and compute the \evlname to assess the information loss. \evlname is an efficient implementation of smatch for documents and also provides inter-sentence coreference subscore (see Table \ref{tab:windows}). To our surprise, a window size as short as 200, loses only 3 points in \smatch and 12 in coreference subscore. Our transition-based parser typically has a target length of up to 3 times the source length. We, therefore, stick to a window size of 300 tokens in our experiments with a 200 tokens overlap. 



\section{Experiments and Results}

We evaluate \parname on AMR 3.0 corpus -- We use its Multi-Sentence \cite{ogorman-etal-2018-amr} test set for document-level evaluations and the standard AMR 3.0 test set for sentence-level evaluation. For training, we use all of AMR 3.0 dataset including the 50k sentences and 242 document annotations. In addition, we parse the CoNLL 2012 coreference dataset \newcite{pradhan-etal-2012-conll} at the sentence level and project coreference information to graph nodes using the parser-generated alignments (similar to \cite{naseem2021docamr}). This results in a silver AMR corpus with gold coreference annotations (1405 documents).
 We also include this silver data in our training set. We used a maximum learning rate of $5e-4$ for the \parname trained with MS-AMR data and $1e-5$ for the full training set, 4000 warm-up steps, a dropout rate of 0.2 and trained up to 120 epochs. For prediction, we used a beam of size 10 to decode the double-annotation part of MS-AMR (dev set) and used \evlname to find top 5 checkpoints and averaged over them. All experiments used a single NVIDIA V100 GPU. We use the publicly available implementation of \cite{zhou-etal-2021-structure}\footnote{\href{https://github.com/IBM/transition-amr-parser}{https://github.com/IBM/transition-amr-parser}}. 

We compare against state-of-the-art document AMR parser of \newcite{naseem2021docamr} -- a pipeline consisting of \newcite{zhou-etal-2021-structure} parser and \newcite{joshi-etal-2020-spanbert} coreference resolution. \newcite{joshi-etal-2020-spanbert} splits input text into non-overlapping windows of length up to 512. We also compare with \cite{zhou-etal-2021-structure} for sentence-level parsing. We report both the standard \smatch \cite{smatch} and the more efficient \evlname. We convert the graphs into \repname\footnote{\href{https://github.com/IBM/docAMR}{https://github.com/IBM/docAMR}} representation before computing the Smatch scores. 

Table \ref{tab:results} shows our main result. \parname, when trained on our full training set, performs on par with the SoTA \newcite{naseem2021docamr} while maintaining the sentence-level performance of the underlying \newcite{zhou-etal-2021-structure}. We also ablate the training data, by removing the CoNLL silver data and the sentence level AMR 3.0 gold data -- which degrades the performance substantially. We also test \parname with various window sizes (Table \ref{tab:windows}) to gauge the impact on performance. It is clear that the use of sliding windows greatly improves the parsing performance. The best setup is the window size of 300 tokens with 200 overlap -- likely because this is the largest size for which the target actions can safely fit within the 1024 length. 


\section{Conclusions}
We propose synchronous source and target side sliding windows for document-level AMR parsing. We show that the left-to-right incremental parser of \cite{zhou-etal-2021-structure} is well suited for the effective implementation of this approach.

\section*{Limitations}
Since we circumvent the problem of long documents using constrained windows, some cross-sentential coreference information might fall through the cracks . We show how much of this information is lost in Table 1. The process of constructing these windows also involves breaking down coreference chains into pairwise edges between contiguous mentions in the chain. In doing so, the first mention of the chain which is often the most informative mention, goes unseen when predicting the next mention in the chain.

\section*{Ethics Statement}
We do not anticipate any ethical concerns regarding our model or use of data. The data we use in this experiment is publicly available and/or from a shared task.

\bibliography{anthology,custom}
\bibliographystyle{acl_natbib}

\appendix



\end{document}